\documentclass{article}

\usepackage[final]{neurips_2022_ml4ps}
\usepackage{graphicx}
\usepackage{amsmath}

\usepackage{natbib}
\usepackage[utf8]{inputenc} 
\usepackage[T1]{fontenc}    
\usepackage{hyperref}       
\usepackage{url}            
\usepackage{booktabs}       
\usepackage{amsfonts}       
\usepackage{nicefrac}       
\usepackage{microtype}      
\usepackage{xcolor}         

\title{Wavelets Beat Monkeys at Adversarial Robustness}

\author{%
  Jingtong Su\\
  Center for Data Science\\
  New York University\\
  \texttt{js12196@nyu.edu} \\
   \And
   Julia Kempe \\
   Center for Data Science and \\
   Courant Institute of Mathematical Sciences\\
   New York University \\
   \texttt{kempe@nyu.edu} \\
}

\begin{document}

\maketitle

\begin{abstract}
Research on improving the robustness of neural networks to adversarial noise -  imperceptible malicious perturbations of the data - has received significant attention. Neural nets struggle to recognize corrupted images that are easily recognized by humans. The currently uncontested state-of-the-art defense to obtain robust deep neural networks is Adversarial Training (AT), but it consumes significantly more resources compared to standard training and trades off accuracy for robustness.
An inspiring recent work \citep{dapello2020simulating} aims to bring neurobiological tools to the question: How can we develop Neural Nets that robustly generalize like human vision? They design a network structure with a neural hidden first layer 
that mimics the primate primary visual cortex (V1), followed by a back-end structure adapted from current CNN vision models. This front-end layer, called VOneBlock, consists of a biologically inspired Gabor Filter Bank with fixed handcrafted "biologically constrained" weights, simple and complex cell non-linearities and a "V1 stochasticity generator" injecting randomness. It seems to achieve non-trivial adversarial robustness on standard vision benchmarks when tested on small perturbations.

Here we revisit this biologically inspired work, which heavily relies on handcrafted tuning of the parameters of the V1 unit based on neural responses derived from experimental records of macaque monkeys. We ask whether a principled parameter-free representation with inspiration from physics is able to achieve the same goal. We discover that the wavelet scattering transform can replace the complex V1-cortex and simple uniform Gaussian noise can take the role of neural stochasticity, to achieve adversarial robustness.
In extensive experiments on the CIFAR-10 benchmark with adaptive adversarial attacks we show that: 1) Robustness of VOneBlock architectures is relatively weak (though non-zero) when the strength of the adversarial attack radius is set to commonly used benchmarks. 2) Replacing the front-end VOneBlock by an off-the-shelf parameter-free Scatternet followed by simple uniform Gaussian noise can achieve much more substantial adversarial robustness without adversarial training. 
Our work shows how physically inspired structures yield new insights into robustness that were previously only thought possible by meticulously mimicking the human cortex. Physics, rather than only neuroscience, can guide us towards more robust neural networks.
\end{abstract}

\section{Introduction}

Deep Neural Networks (DNNs) are shown to be extremely sensitive to test time perturbations \citep{szegedy2013intriguing, goodfellow2014explaining}. Take the object recognition task in the computer vision field as an example; an adversary can perturb the input image by only a few pixels values per pixel point to change the model prediction \citet{su2019one}.

Formally, given a loss function $\mathcal L$, a parametric model $f(\cdot;\theta)$ with parameters $\theta$, for an input sample $x$ with label $y$ an adversary will try to find the worst case nearby input point $x'=x+\delta$:
\begin{equation}
\label{eqa1}
    \delta = \arg\max_{||\delta||\le \epsilon} \mathcal L(f(x+\delta;\theta), y)
\end{equation}
for some notion of $\epsilon$-closeness, in computer vision usually given by either $||\cdot||_2$ or $||\cdot||_\infty$.

\textbf{Defenses.} To defend against such attacks, a rich body of research works has addressed the problem from several viewpoints (\citet{papernot2016distillation, madry2017towards, ilyas2019adversarial}). Among them, Adversarial Training (AT) (\citet{madry2017towards, goodfellow2014explaining}), an iterative procedure that successively optimizes model parameters and computes worst-case augmentations for the training data, has become the gold standard baseline to achieve robustness. However, AT consumes significantly more resources to train, sacrifices test accuracy, and thus is limited in real-world applications (\citet{shafahi2019adversarial, wong2020fast, tsipras2018robustness}). Several works build upon AT to learn robust data embeddings (\citep{pang2020boosting, li2021towards}); however, there are limited works that aim to achieve adversarial robustness without adversarial training, by extracting robust representations from data directly. \citet{yang2020closer} point out that several natural image datasets are separated, and thus imply the existence of robust and accurate classifiers with local Lipschitzness. \citet{Gar+18} propose a spectral-based function that is robust near the training set, and \citet{awasthi2021adversarially} devise a robust PCA algorithm to project data in a principled way. These two methods operate on the training set, and no test time robustness was shown, compared to AT. For NLP tasks, \citep{jones2020robust} propose a robust encoding, by projecting words to a smaller and discrete space where similar inputs share exactly the same encoding.

\textbf{Inspiration from Neuroscience.} Arguably, current state-of-the art object recognition models based on convolutional neural nets (CNN) are inspired by the human visual system, and a series of research work has aimed to infuse computer vision with ideas from neuroscience, trying to more closely align neural networks and human vision (see e.g. \citep{kubilius2019brain, lindsay2018biological, geirhos2021partial}). \citep{dapello2020simulating} are the first to address the robustness problem from this view in a principled structural way. Based on rich prior work on V1-modeling, they devise a systematic architecture, called the VOneBlock, as a preprocessing technique to the inputs of a back-end convolutional neural network. They claim models assisted by their block, collectively called ``V1 models'', can be more difficult to fool, and are even as robust as those trained with state-of-the-art Adversarial Training.

\textbf{Our work.} In this paper, we seek to investigate why and how V1 models function and explore whether insights from physics can help to simplify and extract the underlying principles. To this end, we first revisit the VOne block proposed in \citep{dapello2020simulating} and test it on the CIFAR-10 dataset  \citep{krizhevsky2009learning} with a relatively simple convolutional back-end, using both the gradient-based PGD-attack (\cite{KGB17,Mad+18}) and adaptive attacks from the AutoAttack benchmark (\citep{croce2020reliable}) with standard strength ($\epsilon$=8/255 for $\ell_{\infty}$-attacks at which point most CNNs show $0\%$ robust accuracy).
Surprisingly, we reveal that  V1 models are not as powerful as believed when tested with smaller attack radius as in \cite{dapello2020simulating}. We first show that under fair comparison, the V1 models are only slightly robust ($\sim$ 10\% test robustness), and perform strictly worse than AT. We revisit the ablation study performed in \citep{dapello2020simulating} in standard attack settings to show that when we remove any one of the components of the VOneBlock the model looses any robustness whatsoever, demonstrating that the delicate interplay of these various handcrafted features is necessary.

\textbf{Enter wavelets.} 
The scattering transform was originally introduced in the mathematics literature with follow-up works that appeared in the signal processing and computer science literature. Introduced by \citep{mallat2012group}, it combines wavelet multiscale decompositions with a deep convolutional architecture. 
More recently, it has been used in a number of scientific applications: intermittency in turbulence \citep{bruna2015intermittent}, quantum
chemistry and material science \citep{hirn2017wavelet, eickenberg2018solid, sinz2020wavelet}, plasma physics \citep{glinsky2020quantification}, geography \citep{kavalerov2019three}, astrophysics \citep{allys2019rwst, saydjari2021classification, regaldo2020statistical}, and cosmology \citep{cheng2020new, cheng2021weak} (see \citep{cheng2021quantify}).
The use of the scattering transform for data preprocessing to improve machine-learning tasks has recently emerged in the context of {\em differential privacy}. \citep{tramer2021differentially} show that ScatterNet used as a feature extractor improves upon the privacy-utility trade-off of deep learning.  

\begin{figure}[h]
\centering
\vskip -0.15in
\includegraphics[width=0.9\textwidth]{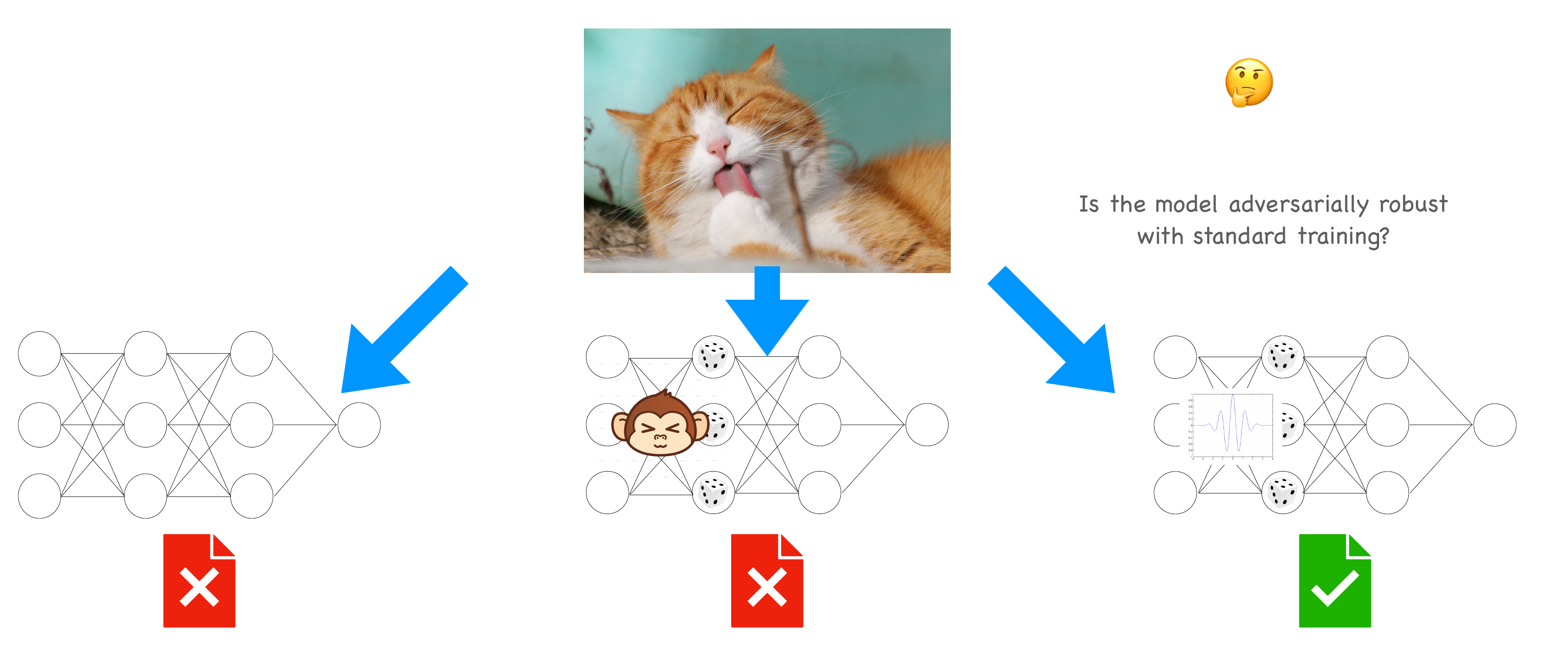}
\caption{Illustration of model construction discussed in our paper. Only our Stochastic ScatterNet model retains genuine robustness. \textbf{Left:} Standard CNNs. \textbf{Middle:} the V1 model. The first layer is replaced with the Gabor Filter Bank (GFB) to approximate empirical primate V1 neural response, which requires massive experimental records to craft the weights, followed by injection of (neuronal) noise. \textbf{Right:} Our proposed Stochastic ScatterNet model. We use a mathematically- and physically-motivated wavelet transform in the first layer, followed by injection of uniform Gaussian noise.}
\label{fig:illustration}
\vskip -0.15in
\end{figure}

We show that we can replace the biologically-inspired VOneBlock by the off-the-shelf ScatterNet \citet{oyallon2015deep,bruna2013invariant}, followed with injection of uniform Gaussian noise. The only one parameter we have tuned is the variance of the noise. We achieve surprising genuine robustness that far surpasses the VOneBlock in this regime. A cartoonish summary of our work is given in Figure \ref{fig:illustration}.

We summarize our contributions as follows:
\begin{enumerate}
    \item We replicate previous studies of the VOneBlock on CIFAR-10 for standard settings of attack strength and find that its robustness, while non-zero, falls below $10\%$ (random guessing), and is considerably lower than for the adversarially-trained model  (Table \ref{tab:V1Baseline}). Previous claims of AT-comparable PGD-robustness of the VOneBlock only hold when weakening the strength of the attacks.
    \item We demonstrate that none of the components of VOne alone induce any robustness. We conclude the VOneBlock as a whole is the key towards slight robustness.
    \item We replace the biologically-inspired VOne component with our Stochastic ScatterNet and show that it achieves significantly larger robustness (Table \ref{tab:ScatterNet}). This opens the route to systematic exploration of the wavelet scattering transform as a preprocessing tool to achieve human-like robust generalization. 
\end{enumerate}

\section{Preliminaries}

\textbf{The V1 model.} We adopt the biologically-inspired VOneBlock as proposed in \citep{dapello2020simulating}, which consists of a \textit{fixed-weight} convolutional layer called the Gabor Filter Bank (GFB) with corresponding nonlinearty, and a stochastic layer called the V1 stochasticity: $\textrm{VOne}(x) = sto(\textrm{GFB}(x))$.

The mathematically parameterized GFB has its parameters tuned to approximate empirical primate V1 neural response data. It convolves the RGB input images with Gabor filters of multiple orientations, sizes, shapes, and spatial frequencies. To instantiate a VOneBlock, we use publicly available code that randomly samples (and then fixes) the values for the GFB parameters according to empirically observed distributions of preferred orientation, peak spatial frequency, and size/shape of receptive fields \citep{de1982orientation, de1982spatial, ringach2002spatial}.

A defining property of neuronal responses is their stochasticity.  In
awake monkeys, the spike train (corresponding to {\em activations}) for each trial is approximately Poisson: the spike count variance is equal to the mean. Thus, the VOne stochasticity - also referred to as {\em neuronal noise} - injects noise into the resulting activations $z_i$ as
\begin{equation}
\label{eq:neuronal_affine}
    sto(z_i) = z_i + \frac{\mathcal N(0,|l(z_i)|)}{a},\ 
    l(z_i) = a\cdot z_i + b.
\end{equation}
The affine transformation $l(z)$ serves to normalize mean activations to correspond to those of a population of primate V1 neurons measured in a 25ms time-window, and $a$ and $b$ are set accordingly (see App. \ref{experiment} for more detail).

In our ablation studies for the VOne block we study several simplifications of the neuronal noise.
\begin{itemize}
    \item V1-Magnitude Gaussian: We remove the affine transformation $l(z)$ (setting $a=1$ and $b=0$). This results in \textbf{magnitude-aware Gaussian noise}   $sto_{MG}(z_i) = z_i + \mathcal N(0, |z_i|)$.
    \item V1-Gaussian: We use uniform \textbf{Gaussian noise} $Gau(z_i) = z_i + \mathcal N(0,\sigma^2) $ setting $\sigma^2$ to a constant. 
    \item V1-None: We remove the injected noise altogether.
\end{itemize}
\section{Experimental Results on CIFAR-10}

Throughout our experiments, we use a simple three-layer CNN of width 64 with Max-Pooling as the back-end model.  For the V1 model we replace its first layer with the VOne block. We use the ScatterNet implementation from \textit{Kymatio} \citet{andreux2020kymatio}. Full experimental details can be found in Appendix \ref{experiment}.
For adversarial robustness, we use FGSM \cite{goodfellow2014explaining} and the standard $\ell_\infty$ Projected Gradient Descent (PGD), with 20 steps, step size $\sigma=2/255$ and budget $\epsilon=8/255$. To check for genuine robustness, we apply the AutoAttack (AA) benchmark \cite{croce2020reliable} with the same strength.
Because we introduce stochasticity, we need to take extra care when attacking our models.  See Appendix \ref{adversarial}.

\paragraph {Robustness of VOne:} Table \ref{tab:V1Baseline} shows test statistics of our V1 models, together with several variations of stochasticity, and the adversarially trained baseline. We find that robustness of V1 models is only slight, though genuine and non-zero, and much inferior to the AT model. This comparison was much more favorable in \citep{dapello2020simulating}, and is manifestly due to {\em weakening the strength of the attack} to $1/16$ of the standard magnitude. However, since the baseline CNN has {\em zero} robustness, we conclude that V1 models garner some robustness benefits. None of the VOne components separately give meaningful robustness, as discussed in Appendix \ref{ablations}.

\begin{table}[h]
    \caption{V1 model v.s. AT baseline CIFAR-10 Test Performance ($\%$). We boldface the \textbf{best} result.}
    \centering
    \begin{tabular}{ccccc}
        \hline
        & & \multicolumn{3}{c}{\textbf{Robust}} \\
        \textbf{Model} & \textbf{Clean} & FGSM & PGD $\ell_\infty$ 20 & AA \\
        \hline
        V1-Neuronal & 58.66 $\pm$ 0.56 & \textbf{18.07 $\pm$ 1.54} & \textbf{9.47 $\pm$ 0.70} & \textbf{27.57 $\pm$ 0.80}  \\
         V1-Magnitude Gaussian & 63.44 $\pm$ 0.44 & 7.14 $\pm$ 0.58 & 0.71 $\pm$ 0.16 & 10.55 $\pm$ 0.87\\
         
        V1-Std-0.35-Gaussian & 63.01 $\pm$ 1.17 & 1.74 $\pm$ 0.32 & 0.01 $\pm$ 0.01 & 7.22 $\pm$ 0.60 \\

        V1-Std-0.20-Gaussian & 63.75 $\pm$ 0.47 & 2.02 $\pm$ 0.20 & 0.00 $\pm$ 0.00 & 5.64 $\pm$ 0.41 \\
         
        V1-Std-0.15-Gaussian & 64.07 $\pm$ 0.46 & 1.91 $\pm$ 0.14 & 0.03 $\pm$ 0.03 & 4.62 $\pm$ 0.13 \\
        
        V1-Std-0.10-Gaussian & 63.86 $\pm$ 1.46 & 1.29 $\pm$ 0.14 & 0.05 $\pm$ 0.04 & 3.96 $\pm$ 0.28 \\  
        
        V1-None & \textbf{64.42 $\pm$ 1.13} & 1.51 $\pm$ 0.51 & 0.93 $\pm$ 0.60 & 0.00 $\pm$ 0.00\\

        \hline
        AT Baseline & 58.07 & 33.94 & 31.49 & 26.18 \\
        \hline
    \end{tabular}
   \vskip -0.1in
    \label{tab:V1Baseline}
\end{table}

\paragraph{Robustness of ScatterNet:} 
Table \ref{tab:ScatterNet} shows robustness of our Stochastic ScatterNet with uniform Gaussian noise of various magnitudes instead of the VOne module. Surprisingly, we observe that the wavelet scattering transform combined with simple uniform Gaussian noise can achieve much higher robustness than biologically-motivated V1-models.

\begin{table}[h]
   \caption{Stochastic ScatterNet Model CIFAR-10 Test Performance ($\%$). We boldface the \textbf{best} result.}
    \centering
    \begin{tabular}{ccccc}
        \hline
        & & \multicolumn{3}{c}{\textbf{Robust}} \\
        \textbf{Stochasticity Type} & \textbf{Clean} & FGSM & PGD $\ell_\infty$ 20 & AA \\
        \hline
        Magnitude Gaussian & 58.55 $\pm$ 0.21 & 16.98 $\pm$ 0.29 & 4.11 $\pm$ 0.14 & 22.90 $\pm$ 0.30  \\
        Std-0.35-Gaussian & 36.02 $\pm$ 0.42 & 23.99 $\pm$ 0.64 & 22.98 $\pm$ 0.26 & 33.52 $\pm$ 0.38 \\
        Std-0.2-Gaussian & 46.12 $\pm$ 0.36 & 26.16 $\pm$ 0.12 & \textbf{24.84 $\pm$ 0.57} & 39.88 $\pm$ 0.52 \\
        Std-0.15-Gaussian & 50.83 $\pm$ 0.42 & \textbf{26.38 $\pm$ 0.44} & 24.03 $\pm$ 0.08 & \textbf{41.43 $\pm$ 0.40} \\
        Std-0.1-Gaussian & 56.27 $\pm$ 0.59 & 24.45 $\pm$ 0.28 & 20.74 $\pm$ 0.24 & 39.75 $\pm$ 0.26 \\
        None & \textbf{76.44 $\pm$ 0.30} & 6.03 $\pm$ 0.20 & 0.04 $\pm$ 0.04 & 0.01 $\pm$ 0.01\\
        \hline
    \end{tabular}
\vskip -0.1in
    \label{tab:ScatterNet}
\end{table}
In particular, we observe that simple uniform Gaussian noise leads to higher robustness than neuronal noise of varying magnitude. Yet, similar to the VOne architecture, stochasticity seems indispensable for robustness.

\section{Conclusion}
In this paper, we show that the biologically-inspired VOneBlock is only slightly robust under standard attack settings, and none of its components can function by itself. Surprisingly, we observe that the wavelet scattering transform combined with simple uniform Gaussian noise can achieve much higher robustness than these well-motivated models which require mounts of experimental primate data to craft. In future work, we aim to explore how our stochastic ScatterNet fares for other dataset and back-end architectures and to gain further theoretical understanding of this surprising robustness.

We advocate to continue to study how bringing insights from physical models can assist vision tasks, in addition to biologically-inspired ones.

\section*{Acknowledgements}
This work was supported in part through the NYU IT High Performance Computing resources, services, and staff expertise. Both authors were supported by the National Science Foundation under NSF Award 1922658.

\bibliography{ref}

\begin{thebibliography}{45}
\providecommand{\natexlab}[1]{#1}
\providecommand{\url}[1]{\texttt{#1}}
\expandafter\ifx\csname urlstyle\endcsname\relax
  \providecommand{\doi}[1]{doi: #1}\else
  \providecommand{\doi}{doi: \begingroup \urlstyle{rm}\Url}\fi

\bibitem[Dapello et~al.(2020)Dapello, Marques, Schrimpf, Geiger, Cox, and
  DiCarlo]{dapello2020simulating}
Joel Dapello, Tiago Marques, Martin Schrimpf, Franziska Geiger, David Cox, and
  James~J DiCarlo.
\newblock Simulating a primary visual cortex at the front of cnns improves
  robustness to image perturbations.
\newblock \emph{Advances in Neural Information Processing Systems},
  33:\penalty0 13073--13087, 2020.

\bibitem[Szegedy et~al.(2013)Szegedy, Zaremba, Sutskever, Bruna, Erhan,
  Goodfellow, and Fergus]{szegedy2013intriguing}
Christian Szegedy, Wojciech Zaremba, Ilya Sutskever, Joan Bruna, Dumitru Erhan,
  Ian Goodfellow, and Rob Fergus.
\newblock Intriguing properties of neural networks.
\newblock \emph{arXiv preprint arXiv:1312.6199}, 2013.

\bibitem[Goodfellow et~al.(2014)Goodfellow, Shlens, and
  Szegedy]{goodfellow2014explaining}
Ian~J Goodfellow, Jonathon Shlens, and Christian Szegedy.
\newblock Explaining and harnessing adversarial examples.
\newblock \emph{arXiv preprint arXiv:1412.6572}, 2014.

\bibitem[Su et~al.(2019)Su, Vargas, and Sakurai]{su2019one}
Jiawei Su, Danilo~Vasconcellos Vargas, and Kouichi Sakurai.
\newblock One pixel attack for fooling deep neural networks.
\newblock \emph{IEEE Transactions on Evolutionary Computation}, 23\penalty0
  (5):\penalty0 828--841, 2019.

\bibitem[Papernot et~al.(2016)Papernot, McDaniel, Wu, Jha, and
  Swami]{papernot2016distillation}
Nicolas Papernot, Patrick McDaniel, Xi~Wu, Somesh Jha, and Ananthram Swami.
\newblock Distillation as a defense to adversarial perturbations against deep
  neural networks.
\newblock In \emph{2016 IEEE symposium on security and privacy (SP)}, pages
  582--597. IEEE, 2016.

\bibitem[Madry et~al.(2017)Madry, Makelov, Schmidt, Tsipras, and
  Vladu]{madry2017towards}
Aleksander Madry, Aleksandar Makelov, Ludwig Schmidt, Dimitris Tsipras, and
  Adrian Vladu.
\newblock Towards deep learning models resistant to adversarial attacks.
\newblock \emph{arXiv preprint arXiv:1706.06083}, 2017.

\bibitem[Ilyas et~al.(2019)Ilyas, Santurkar, Tsipras, Engstrom, Tran, and
  Madry]{ilyas2019adversarial}
Andrew Ilyas, Shibani Santurkar, Dimitris Tsipras, Logan Engstrom, Brandon
  Tran, and Aleksander Madry.
\newblock Adversarial examples are not bugs, they are features.
\newblock \emph{Advances in neural information processing systems}, 32, 2019.

\bibitem[Shafahi et~al.(2019)Shafahi, Najibi, Ghiasi, Xu, Dickerson, Studer,
  Davis, Taylor, and Goldstein]{shafahi2019adversarial}
Ali Shafahi, Mahyar Najibi, Mohammad~Amin Ghiasi, Zheng Xu, John Dickerson,
  Christoph Studer, Larry~S Davis, Gavin Taylor, and Tom Goldstein.
\newblock Adversarial training for free!
\newblock \emph{Advances in Neural Information Processing Systems}, 32, 2019.

\bibitem[Wong et~al.(2020)Wong, Rice, and Kolter]{wong2020fast}
Eric Wong, Leslie Rice, and J~Zico Kolter.
\newblock Fast is better than free: Revisiting adversarial training.
\newblock \emph{arXiv preprint arXiv:2001.03994}, 2020.

\bibitem[Tsipras et~al.(2018)Tsipras, Santurkar, Engstrom, Turner, and
  Madry]{tsipras2018robustness}
Dimitris Tsipras, Shibani Santurkar, Logan Engstrom, Alexander Turner, and
  Aleksander Madry.
\newblock Robustness may be at odds with accuracy.
\newblock \emph{arXiv preprint arXiv:1805.12152}, 2018.

\bibitem[Pang et~al.(2020)Pang, Yang, Dong, Xu, Zhu, and Su]{pang2020boosting}
Tianyu Pang, Xiao Yang, Yinpeng Dong, Kun Xu, Jun Zhu, and Hang Su.
\newblock Boosting adversarial training with hypersphere embedding.
\newblock \emph{Advances in Neural Information Processing Systems},
  33:\penalty0 7779--7792, 2020.

\bibitem[Li et~al.(2021)Li, Min, Lee, Yu, Kruus, Wang, and
  Hsieh]{li2021towards}
Yao Li, Martin~Renqiang Min, Thomas Lee, Wenchao Yu, Erik Kruus, Wei Wang, and
  Cho-Jui Hsieh.
\newblock Towards robustness of deep neural networks via regularization.
\newblock In \emph{Proceedings of the IEEE/CVF International Conference on
  Computer Vision}, pages 7496--7505, 2021.

\bibitem[Yang et~al.(2020)Yang, Rashtchian, Zhang, Salakhutdinov, and
  Chaudhuri]{yang2020closer}
Yao-Yuan Yang, Cyrus Rashtchian, Hongyang Zhang, Russ~R Salakhutdinov, and
  Kamalika Chaudhuri.
\newblock A closer look at accuracy vs. robustness.
\newblock \emph{Advances in neural information processing systems},
  33:\penalty0 8588--8601, 2020.

\bibitem[Garg et~al.(2018)Garg, Sharan, Zhang, and Valiant]{Gar+18}
Shivam Garg, Vatsal Sharan, Brian~Hu Zhang, and Gregory Valiant.
\newblock A spectral view of adversarially robust features.
\newblock In Samy Bengio, Hanna~M. Wallach, Hugo Larochelle, Kristen Grauman,
  Nicol{\`{o}} Cesa{-}Bianchi, and Roman Garnett, editors, \emph{Advances in
  Neural Information Processing Systems 31: Annual Conference on Neural
  Information Processing Systems 2018, NeurIPS 2018, December 3-8, 2018,
  Montr{\'{e}}al, Canada}, pages 10159--10169, 2018.

\bibitem[Awasthi et~al.(2021)Awasthi, Chatziafratis, Chen, and
  Vijayaraghavan]{awasthi2021adversarially}
Pranjal Awasthi, Vaggos Chatziafratis, Xue Chen, and Aravindan Vijayaraghavan.
\newblock Adversarially robust low dimensional representations.
\newblock In \emph{Conference on Learning Theory}, pages 237--325. PMLR, 2021.

\bibitem[Jones et~al.(2020)Jones, Jia, Raghunathan, and Liang]{jones2020robust}
Erik Jones, Robin Jia, Aditi Raghunathan, and Percy Liang.
\newblock Robust encodings: A framework for combating adversarial typos.
\newblock \emph{arXiv preprint arXiv:2005.01229}, 2020.

\bibitem[Kubilius et~al.(2019)Kubilius, Schrimpf, Kar, Rajalingham, Hong,
  Majaj, Issa, Bashivan, Prescott-Roy, Schmidt, et~al.]{kubilius2019brain}
Jonas Kubilius, Martin Schrimpf, Kohitij Kar, Rishi Rajalingham, Ha~Hong, Najib
  Majaj, Elias Issa, Pouya Bashivan, Jonathan Prescott-Roy, Kailyn Schmidt,
  et~al.
\newblock Brain-like object recognition with high-performing shallow recurrent
  anns.
\newblock \emph{Advances in neural information processing systems}, 32, 2019.

\bibitem[Lindsay and Miller(2018)]{lindsay2018biological}
Grace~W Lindsay and Kenneth~D Miller.
\newblock How biological attention mechanisms improve task performance in a
  large-scale visual system model.
\newblock \emph{ELife}, 7:\penalty0 e38105, 2018.

\bibitem[Geirhos et~al.(2021)Geirhos, Narayanappa, Mitzkus, Thieringer, Bethge,
  Wichmann, and Brendel]{geirhos2021partial}
Robert Geirhos, Kantharaju Narayanappa, Benjamin Mitzkus, Tizian Thieringer,
  Matthias Bethge, Felix~A Wichmann, and Wieland Brendel.
\newblock Partial success in closing the gap between human and machine vision.
\newblock \emph{Advances in Neural Information Processing Systems},
  34:\penalty0 23885--23899, 2021.

\bibitem[Krizhevsky et~al.(2009)Krizhevsky, Hinton,
  et~al.]{krizhevsky2009learning}
Alex Krizhevsky, Geoffrey Hinton, et~al.
\newblock Learning multiple layers of features from tiny images.
\newblock 2009.

\bibitem[Kurakin et~al.(2017)Kurakin, Goodfellow, and Bengio]{KGB17}
Alexey Kurakin, Ian~J. Goodfellow, and Samy Bengio.
\newblock Adversarial examples in the physical world.
\newblock In \emph{5th International Conference on Learning Representations,
  {ICLR} 2017, Toulon, France, April 24-26, 2017, Workshop Track Proceedings},
  2017.

\bibitem[Madry et~al.(2018)Madry, Makelov, Schmidt, Tsipras, and Vladu]{Mad+18}
Aleksander Madry, Aleksandar Makelov, Ludwig Schmidt, Dimitris Tsipras, and
  Adrian Vladu.
\newblock {Towards Deep Learning Models Resistant to Adversarial Attacks}.
\newblock In \emph{International Conference on Learning Representations}, 2018.

\bibitem[Croce and Hein(2020)]{croce2020reliable}
Francesco Croce and Matthias Hein.
\newblock Reliable evaluation of adversarial robustness with an ensemble of
  diverse parameter-free attacks.
\newblock In \emph{ICML}, 2020.

\bibitem[Mallat(2012)]{mallat2012group}
St{\'e}phane Mallat.
\newblock Group invariant scattering.
\newblock \emph{Communications on Pure and Applied Mathematics}, 65\penalty0
  (10):\penalty0 1331--1398, 2012.

\bibitem[Bruna et~al.(2015)Bruna, Mallat, Bacry, and
  Muzy]{bruna2015intermittent}
Joan Bruna, St{\'e}phane Mallat, Emmanuel Bacry, and Jean-Fran{\c{c}}ois Muzy.
\newblock Intermittent process analysis with scattering moments.
\newblock \emph{The Annals of Statistics}, 43\penalty0 (1):\penalty0 323--351,
  2015.

\bibitem[Hirn et~al.(2017)Hirn, Mallat, and Poilvert]{hirn2017wavelet}
Matthew Hirn, St{\'e}phane Mallat, and Nicolas Poilvert.
\newblock Wavelet scattering regression of quantum chemical energies.
\newblock \emph{Multiscale Modeling \& Simulation}, 15\penalty0 (2):\penalty0
  827--863, 2017.

\bibitem[Eickenberg et~al.(2018)Eickenberg, Exarchakis, Hirn, Mallat, and
  Thiry]{eickenberg2018solid}
Michael Eickenberg, Georgios Exarchakis, Matthew Hirn, St{\'e}phane Mallat, and
  Louis Thiry.
\newblock Solid harmonic wavelet scattering for predictions of molecule
  properties.
\newblock \emph{The Journal of chemical physics}, 148\penalty0 (24):\penalty0
  241732, 2018.

\bibitem[Sinz et~al.(2020)Sinz, Swift, Brumwell, Liu, Kim, Qi, and
  Hirn]{sinz2020wavelet}
Paul Sinz, Michael~W Swift, Xavier Brumwell, Jialin Liu, Kwang~Jin Kim, Yue Qi,
  and Matthew Hirn.
\newblock Wavelet scattering networks for atomistic systems with extrapolation
  of material properties.
\newblock \emph{The Journal of Chemical Physics}, 153\penalty0 (8):\penalty0
  084109, 2020.

\bibitem[Glinsky et~al.(2020)Glinsky, Moore, Lewis, Weis, Jennings, Ampleford,
  Knapp, Harding, Gomez, and Harvey-Thompson]{glinsky2020quantification}
Michael~E Glinsky, Thomas~W Moore, William~E Lewis, Matthew~R Weis,
  Christopher~A Jennings, David~J Ampleford, Patrick~F Knapp, Eric~C Harding,
  Matthew~R Gomez, and Adam~J Harvey-Thompson.
\newblock Quantification of maglif morphology using the mallat scattering
  transformation.
\newblock \emph{Physics of Plasmas}, 27\penalty0 (11):\penalty0 112703, 2020.

\bibitem[Kavalerov et~al.(2019)Kavalerov, Li, Czaja, and
  Chellappa]{kavalerov2019three}
Ilya Kavalerov, Weilin Li, Wojciech Czaja, and Rama Chellappa.
\newblock Three-dimensional fourier scattering transform and classification of
  hyperspectral images.
\newblock \emph{arXiv preprint arXiv:1906.06804}, 2019.

\bibitem[Allys et~al.(2019)Allys, Levrier, Zhang, Colling,
  Regaldo-Saint~Blancard, Boulanger, Hennebelle, and Mallat]{allys2019rwst}
Erwan Allys, F~Levrier, S~Zhang, C~Colling, B~Regaldo-Saint~Blancard,
  F~Boulanger, P~Hennebelle, and S~Mallat.
\newblock The rwst, a comprehensive statistical description of the non-gaussian
  structures in the ism.
\newblock \emph{Astronomy \& Astrophysics}, 629:\penalty0 A115, 2019.

\bibitem[Saydjari et~al.(2021)Saydjari, Portillo, Slepian, Kahraman, Burkhart,
  and Finkbeiner]{saydjari2021classification}
Andrew~K Saydjari, Stephen~KN Portillo, Zachary Slepian, Sule Kahraman,
  Blakesley Burkhart, and Douglas~P Finkbeiner.
\newblock Classification of magnetohydrodynamic simulations using wavelet
  scattering transforms.
\newblock \emph{The Astrophysical Journal}, 910\penalty0 (2):\penalty0 122,
  2021.

\bibitem[Regaldo-Saint~Blancard et~al.(2020)Regaldo-Saint~Blancard, Levrier,
  Allys, Bellomi, and Boulanger]{regaldo2020statistical}
Bruno Regaldo-Saint~Blancard, Fran{\c{c}}ois Levrier, Erwan Allys, Elena
  Bellomi, and Fran{\c{c}}ois Boulanger.
\newblock Statistical description of dust polarized emission from the diffuse
  interstellar medium-a rwst approach.
\newblock \emph{Astronomy \& Astrophysics}, 642:\penalty0 A217, 2020.

\bibitem[Cheng et~al.(2020)Cheng, Ting, M{\'e}nard, and Bruna]{cheng2020new}
Sihao Cheng, Yuan-Sen Ting, Brice M{\'e}nard, and Joan Bruna.
\newblock A new approach to observational cosmology using the scattering
  transform.
\newblock \emph{Monthly Notices of the Royal Astronomical Society},
  499\penalty0 (4):\penalty0 5902--5914, 2020.

\bibitem[Cheng and M{\'e}nard(2021{\natexlab{a}})]{cheng2021weak}
Sihao Cheng and Brice M{\'e}nard.
\newblock Weak lensing scattering transform: dark energy and neutrino mass
  sensitivity.
\newblock \emph{Monthly Notices of the Royal Astronomical Society},
  507\penalty0 (1):\penalty0 1012--1020, 2021{\natexlab{a}}.

\bibitem[Cheng and M{\'e}nard(2021{\natexlab{b}})]{cheng2021quantify}
Sihao Cheng and Brice M{\'e}nard.
\newblock How to quantify fields or textures? a guide to the scattering
  transform.
\newblock \emph{arXiv preprint arXiv:2112.01288}, 2021{\natexlab{b}}.

\bibitem[Tramer and Boneh(2021)]{tramer2021differentially}
Florian Tramer and Dan Boneh.
\newblock Differentially private learning needs better features (or much more
  data).
\newblock In \emph{International Conference on Learning Representations}, 2021.

\bibitem[Oyallon and Mallat(2015)]{oyallon2015deep}
Edouard Oyallon and St{\'e}phane Mallat.
\newblock Deep roto-translation scattering for object classification.
\newblock In \emph{Proceedings of the IEEE Conference on Computer Vision and
  Pattern Recognition}, pages 2865--2873, 2015.

\bibitem[Bruna and Mallat(2013)]{bruna2013invariant}
Joan Bruna and St{\'e}phane Mallat.
\newblock Invariant scattering convolution networks.
\newblock \emph{IEEE transactions on pattern analysis and machine
  intelligence}, 35\penalty0 (8):\penalty0 1872--1886, 2013.

\bibitem[De~Valois et~al.(1982{\natexlab{a}})De~Valois, Yund, and
  Hepler]{de1982orientation}
Russell~L De~Valois, E~William Yund, and Norva Hepler.
\newblock The orientation and direction selectivity of cells in macaque visual
  cortex.
\newblock \emph{Vision research}, 22\penalty0 (5):\penalty0 531--544,
  1982{\natexlab{a}}.

\bibitem[De~Valois et~al.(1982{\natexlab{b}})De~Valois, Albrecht, and
  Thorell]{de1982spatial}
Russell~L De~Valois, Duane~G Albrecht, and Lisa~G Thorell.
\newblock Spatial frequency selectivity of cells in macaque visual cortex.
\newblock \emph{Vision research}, 22\penalty0 (5):\penalty0 545--559,
  1982{\natexlab{b}}.

\bibitem[Ringach(2002)]{ringach2002spatial}
Dario~L Ringach.
\newblock Spatial structure and symmetry of simple-cell receptive fields in
  macaque primary visual cortex.
\newblock \emph{Journal of neurophysiology}, 2002.

\bibitem[Andreux et~al.(2020)Andreux, Angles, Exarchakis, Leonarduzzi,
  Rochette, Thiry, Zarka, Mallat, And{\'e}n, Belilovsky,
  et~al.]{andreux2020kymatio}
Mathieu Andreux, Tom{\'a}s Angles, Georgios Exarchakis, Roberto Leonarduzzi,
  Gaspar Rochette, Louis Thiry, John Zarka, St{\'e}phane Mallat, Joakim
  And{\'e}n, Eugene Belilovsky, et~al.
\newblock Kymatio: Scattering transforms in python.
\newblock \emph{J. Mach. Learn. Res.}, 21\penalty0 (60):\penalty0 1--6, 2020.

\bibitem[Kingma and Ba(2014)]{kingma2014adam}
Diederik~P Kingma and Jimmy Ba.
\newblock Adam: A method for stochastic optimization.
\newblock \emph{arXiv preprint arXiv:1412.6980}, 2014.

\bibitem[Athalye et~al.(2018)Athalye, Carlini, and
  Wagner]{athalye2018obfuscated}
Anish Athalye, Nicholas Carlini, and David Wagner.
\newblock Obfuscated gradients give a false sense of security: Circumventing
  defenses to adversarial examples.
\newblock In \emph{International conference on machine learning}, pages
  274--283. PMLR, 2018.

\end{thebibliography}

 \appendix


\section{Experimental Details}\label{experiment}

\subsection{Training Strategy}
Throughout our paper, we adopt the Adam optimizer \citet{kingma2014adam} to train the V1 models and the Stochastic ScatterNet models. The initial learning rate is fixed to 1e-3, and the number of epochs is fixed to 200. For AT baseline, we train the network with SGD, initial learning rate 1e-1, and decay the learning rate by 10 at the 100- and 150-th epoch. We run each experiment for 3 times, and report the average and standard deviation.

\subsection{Model Structure}
For the AT baseline, we adopt a simple 3-layer convolutional structure, with Max-Pooling and a linear layer in the end. All convolutional layers set the number of output channels to 64, and stride is set to 1. We replace the first layer with a V1 module, or a ScatterNet transform.

For the V1 block\footnote{We adopt the code from \url{https://github.com/dicarlolab/vonenet}}, we use its default setting, consisting of a 512x32x32 feature map. We set the number of input channels of the convolutional layer behind to 512 to fit this module. For the hyperparameters in Eq.~\eqref{eq:neuronal_affine}, we take the default setting provided by the code of \citep{dapello2020simulating} that imitates the mean stimulus response and spontaneous activity: $a=0.35$ and $b=0.07$.

For the ScatterNet, the output shape is 243x8x8 by default. We set the number of input channels of the convolutional layer behind to 243 to fit this module. We also cancel the Max-Pooling after the third layer, to keep the representation dimension at least at the standard, 64x4x4.

A detailed description is given in Table \ref{tab:structure}.

\begin{table}[h]
    \caption{Model illustration in our paper. The dimension of inputs/features and the layer structures are described top-down. We adopt the default setting of the V1 block and the ScatterNet module, leading to the number of output channels of 512 and 243. We adjust the 2nd and 3rd layer of the CNN to fit these feature shapes.}
    \centering
        \resizebox{0.7\textwidth}{!}{
    \begin{tabular}{|c|c|c|}
        \hline
        \textbf{Simple CNN} & \textbf{V1 Model} & \textbf{Stochastic ScatterNet}\\
        \hline
        3x32x32   & 3x32x32    & 3x32x32 \\
        \hline
        Conv + MaxPool & V1 Module + Noise & ScatterNet + Noise\\
        \hline
        64x16x16 & 512x32x32 & 243x8x8\\
        \hline
        Conv + MaxPool  & Conv + MaxPool & Conv + MaxPool\\
        \hline
        64x8x8 & 64x16x16 & 64x4x4\\
        \hline
        Conv + MaxPool & Conv + MaxPool & Conv\\
        \hline
        64x4x4 & 64x8x8 & 64x4x4\\
        \hline
        Linear & Linear & Linear\\
        \hline
        10 & 10 & 10\\
        \hline
    \end{tabular}}
        \vskip 0.1in
    \label{tab:structure}

\end{table}

\section{Adversarial Attacks}\label{adversarial}

Categorized by having access to the model parameters or not, the attack methods that try to solve Eq. \ref{eqa1} can be divided into white-box and black-box, respectively. The most popular attack baseline is a white-box attack, implemented by multi-step Projected Gradient Descent (PGD) \citep{Mad+18, KGB17}. Here the vector norm $||\cdot||$ is usually taken as $||\cdot||_\infty$ or $||\cdot||_2$, and we choose $||\cdot||_\infty$. We calculate the gradient iteratively
\begin{equation}
    \nabla_x \mathcal L(f(x+\delta;\theta),y)
\end{equation}
and update the adversarial example with the steepest direction according to the norm constraint:
\begin{equation}
   x' \leftarrow \Pi_{\mathcal B} \{x' + \alpha\cdot \mathrm{sign} [\nabla_x \mathcal L(f(x+\delta;\theta),y)]\}.
\end{equation}
$\Pi_\mathcal B$ denotes the projection step, to ensure the generated adversarial example in each loop satisfies a certain norm-based constraint, \textit{i.e.,} $\delta = ||x' - x|| \le \epsilon$ for some $\epsilon$. When the number of iterations is 1 this attack is called FGSM \citep{goodfellow2014explaining}.

To combat the effects of noise in the gradients, we adapted our attack such that at every PGD iteration, we take 10 gradient samples and move in the average direction
to increase the loss, following \cite{athalye2018obfuscated}.

We also evaluate robustness against the adaptive suite of the AutoAttack benchmark \citep{croce2020reliable}, which contains both black-box and white-box attacks and is considered a minimal set of attacks to establish genuine robustness. We use the AA  ``rand'' version option to include a gradient averaging mechanism, to produce reliable perturbations for models with stochasticity.

\section{None of the Components of VOne alone Induce Robustness}\label{ablations}

In this section, we perform ablation studies to understand the underlying behaviour of the V1 model, and in particular how each of its elements help the model gain robustness. 
We divide this examination into three main parts. We check the utility of: 1) the Gabor Filter Bank, 2) the stochasticity, namely the neuronal/Gaussian noise, and 3) the affine transformation used with the neuronal noise.

\textbf{Bio-Inspired Gabor Filter Bank Alone Does not Yield Robustness.}
First, we show the Gabor Filter Bank itself cannot extract features that are robust against adversarial perturbations.  
We simply turn off the stochasticity in the VOneBlock, and leave the GFB alone with the trainable back-end, training the model with {\em no noise}. The penultimate row of Table \ref{tab:V1Baseline} shows the test-robustness: V1-None has neither PGD robustness nor AA robustness. The fact that stochasticity is indispensable matches the observations in \citep{dapello2020simulating}.

\textbf{Stochasticity Alone is Insufficient.}
In order to evaluate the utility of stochasticity in a fair way, we remove the Gabor Filter Bank and instead add a {\em fixed-weight, randomly-initialized convolutional layer}. In other words, we replace the GFB whose weights are  drawn from hand-crafted distributions and then fixed, with randomly drawn convolutional weights that are fixed, and retain the stochasticity alone. 
Table \ref{tab:V1withConv} shows the test results of this ``random feature" stochastic model. We see that with stochasticity alone, the models have a little FGSM robustness, and slightly better AA robustness but failing PGD-robustness. Thus, we establish that stochasticity alone not sufficiently useful for robustness, at marginally at best.
\begin{table}[h]
\caption{(Fixed-weights) Random Feature Stochastic Model CIFAR-10 Test Performance ($\%$). We boldface the \textbf{best} result.}
    \centering
    \begin{tabular}{ccccc}
        \hline
        & & \multicolumn{3}{c}{\textbf{Robust}} \\
        \textbf{Stochasticity Type} & \textbf{Clean} & FGSM & PGD $\ell_\infty$ 20 & AA \\
        \hline
        Neuronal & 68.26 $\pm$ 0.52 & \textbf{5.81 $\pm$ 0.31} & 0.07 $\pm$ 0.02 & 11.26 $\pm$ 0.44 \\
        Magnitude Gaussian & \textbf{69.67 $\pm$ 0.96} & 4.74 $\pm$ 0.32 & 0.03 $\pm$ 0.03 & \textbf{14.03 $\pm$ 0.20} \\
        Std-0.35-Gaussian & 64.53 $\pm$ 0.72 & 2.55 $\pm$ 0.05 & 0.04 $\pm$ 0.01 & 5.35 $\pm$ 0.08 \\
         None & 68.66 $\pm$ 0.30 & 2.73 $\pm$ 0.50 & \textbf{2.40 $\pm$ 0.46} & 0.00 $\pm$ 0.00\\
         \hline
    \end{tabular}
    \label{tab:V1withConv}
\end{table}

\textbf{Varying the stochasticity.} To understand whether the biologically-inspired neuronal noise is indispensible for robustness of the VOne block, or whether other, more generic forms of stochasticity would suffice, we replace Neuronal noise with its handcrafted affine transformation by either Magnitude Gaussian (effectively removing the affine transformation) or standard Gaussians of varying magnitude. Results shown in Table \ref{tab:V1Baseline} show that both the affine transformation and the magnitude-dependent variance of the noise seem indispensable. While removing the affine transformation retains at least a modicum of AA-robustness, it fails against PGD attacks. Making the Gaussians of uniform variance further degrades performance and leads to vanishing robustness.

Interestingly, for ScatterNets a similar variation of the stochastic noise, shown in Table \ref{tab:ScatterNet}, has a different effect. Here, uniform Gaussian noise yields much higher robustness than Magnitude Gaussian. We also see that stochasticity is important to achieve robustness using the wavelet scattering transform.


\end{document}